\documentclass[10pt,twocolumn,letterpaper]{article}
\pdfoutput=1
\usepackage[cvprfinal,pagenumbers]{cvpr} 

\usepackage{graphicx}
\usepackage{amsmath}
\usepackage{amssymb}
\usepackage{booktabs}
\usepackage{mathtools}
\usepackage{iac_pkg}
\usepackage{lipsum}
\usepackage{amssymb}
\usepackage{pifont}
\usepackage{multirow}
\usepackage{tabularx, booktabs}
\usepackage[htt]{hyphenat}
\usepackage{color}
\usepackage{xspace}
\usepackage{cite}
\usepackage{overpic}
\usepackage{arydshln}
\usepackage{csquotes}
\usepackage{subcaption}
\usepackage{xcolor}
\usepackage{array} 
\definecolor{citecolor}{RGB}{34,139,34}
\definecolor{lightred}{RGB}{255,100,100}
\definecolor{cell_bisque}{rgb}{1.0, 0.89, 0.77}
\definecolor{cell_blond}{rgb}{0.98, 0.94, 0.75}
\definecolor{cell_blue}{RGB}{155, 187, 228}
\definecolor{princetonorange}{rgb}{1.0, 0.56, 0.0}
\definecolor{pinkpearl}{rgb}{0.91, 0.67, 0.81}
\definecolor{mossgreen}{rgb}{0.68, 0.87, 0.68}

\newcommand{\Section}[1]{\vspace{-1mm} \section{#1} \vspace{-0mm}}
\newcommand{\SubSection}[1]{\vspace{-1mm} \subsection{#1} \vspace{-0mm}}

\usepackage{hyperref}
\hypersetup{pagebackref=true,breaklinks=true,letterpaper=true,colorlinks,bookmarks=false,citecolor=cyan}

\usepackage[capitalize]{cleveref}

\begin{document}

\title{CLIPC8: Face liveness detection algorithm based on image-text pairs and contrastive learning}

\author{Xu Liu$^{1}$$^,$$^{2}$, Shu Zhou$^{2}$$^,$$^{3}$, Yurong Song$^{1}$, Wenzhe Luo$^{2}$, Xin Zhang$^{2}$\\
$^{1}$ Nanjing University of Posts and Telecommunications  $^{2}$  Bank of Jiangsu $^{3}$ Nanjing University \\
{\tt\small liuxu@jsbchina.cn,} {\tt\small shuzhou@smail.nju.edu.cn,} {\tt\small songyr@njupt.edu.cn,}\\
{\tt\small1577124196@qq.com,} {\tt\small 13861673709@163.com}\
}

\maketitle

\begin{abstract}
Face recognition technology is widely used in the financial field, and various types of liveness attack behaviors need to be addressed. Existing liveness detection algorithms are trained on specific training datasets and tested on testing datasets, but their performance and robustness in transferring to unseen datasets are relatively poor. To tackle this issue, we propose a face liveness detection method based on image-text pairs and contrastive learning, dividing liveness attack problems in the financial field into eight categories and using text information to describe the images of these eight types of attacks. The text encoder and image encoder are used to extract feature vector representations for the classification description text and face images, respectively. By maximizing the similarity of positive samples and minimizing the similarity of negative samples, the model learns shared representations between images and texts. The proposed method is capable of effectively detecting specific liveness attack behaviors in certain scenarios, such as those occurring in dark environments or involving the tampering of ID card photos. Additionally, it is also effective in detecting traditional liveness attack methods, such as printing photo attacks and screen remake attacks. The zero-shot capabilities of face liveness detection on five public datasets, including NUAA, CASIA-FASD, Replay-Attack, OULU-NPU and MSU-MFSD also reaches the level of commercial algorithms. The detection capability of proposed algorithm was verified on 5 types of testing datasets, and the results show that the method outperformed commercial algorithms, and the detection rates reached 100\% on multiple datasets. Demonstrating the effectiveness and robustness of introducing image-text pairs and contrastive learning into liveness detection tasks as proposed in this paper. Our source code can be found: \href{https://github.com/reilxlx/AntiSpoofing-Contrastive-learning}{github.com/reilxlx/AntiSpoofing-Contrastive-learning}.
\end{abstract}
\vspace{-5mm}
\Section{Introduction}\label{sec:intro}
Facial recognition, as a convenient and easy-to-use biometric technology, is widely used in identity authentication scenarios. With the development of digital transformation, many traditional offline businesses are gradually shifting online. In the financial sector, facial recognition is often used as a key step in password logins and identity verification. To ensure the security of the facial recognition process, liveness detection is typically integrated into the face capture stage to perform anti-spoofing checks. In recent years, the black industry in the financial sector has frequently attacked facial recognition systems. Perpetrators obtain citizens' identities and their facial images and video information through illegal means, and use methods such as photo replay, printing, AI synthesis, video injection, and adversarial glasses to bypass facial recognition systems. These attacks can be divided into two major categories: presentation attacks and injection attacks. Presentation attacks include printing facial photos on different types of paper (such as black-and-white prints, color prints, sulfuric acid paper, matte paper, copperplate paper, etc.), displaying facial photos or videos through mobile phones or computer screens, using identify card photos, online verification photos, personal ID photos, 3D masks, or T-shaped masks. Injection attacks involve preparing facial videos in advance (including motion sequences or silent videos), or using technologies like deepfakes, 3D synthesis, and Photoshop to create dynamic videos from photos, and then operating them on rooted phones or emulators. Compared to injection attacks, presentation attacks have a lower threshold and a higher rate of reuse. Attackers only need to obtain a person's life photo or ID photo to implement facial fraud. Additionally, the development of image-generated video technology has also reduced the difficulty of implementing face attacks. In facial security defense, liveness detection, as a key component of facial recognition systems, becomes increasingly important. The application of facial recognition technology in the financial sector often involves cooperation with companies such as SenseTime, Megvii, CloudWalk, Alibaba, Baidu, and Tencent, using the commercial privatized liveness detection algorithms provided by these companies for photo liveness detection. However, this method is limited by the frequency of algorithm updates and cannot obtain the latest attack image data from the financial industry in real time, thus making it difficult to train and update versions promptly to effectively detect abnormal images under specific attack methods.

Currently, liveness detection algorithms are trained and tested on specific datasets and have not been comprehensively validated in complex real-world scenarios~\cite{01yang2019face}, thus having limited generalization capabilities on untrained datasets~\cite{02shao2019multi}. This paper, from the perspective of practical application in the financial sector, conducts detection rate analysis of the privatized photo liveness detection algorithms from SenseTime, Megvii, Tencent, and CloudWalk. Using public face liveness detection datasets NUAA~\cite{03tan2010face}, CASIA-FASD~\cite{04zhang2012face}, Replay-Attack~\cite{05chingovska2012effectiveness}, OULU-NPU~\cite{06boulkenafet2017oulu}, MSU-MFSD~\cite{07wen2015face}, and our own multi-class, multi-scenario dataset CLASS-8, we evaluate the performance of these companies' single-image liveness detection algorithms and analyze their performance on six types of datasets. Experimental results show that these four algorithms did not achieve outstanding results on the six datasets, especially weak in the attack images of OULU-NPU. CloudWalk's high false detection rate indicates that these companies' liveness detection algorithms have weak transfer capabilities across different datasets.

Therefore, in response to the low detection success rate of these four commercial privatized liveness detection algorithms on five public liveness detection datasets and the CLASS-8 dataset under specific scenarios, combined with the limitation of commercial companies not being able to timely grasp first-hand attack images, this paper proposes a simple and easy-to-train liveness detection algorithm tailored for the financial sector. This algorithm, based on the concept of multimodal image-text pairs and contrastive learning, aims to use the correlation between image-text pairs to extract the common and unique characteristics of attack images, using textual descriptions to improve the detection rate of single-image liveness detection algorithms. Our proposed CLIPC8 algorithm has strong transfer capabilities and its zero-shot ability in some scenarios of the five datasets NUAA, CASIA-FASD, Replay-Attack, OULU-NPU, MSU-MFSD is not inferior to the four commercial liveness detection algorithms. We believe that in addition to using commercial algorithms in the financial sector, it is also necessary to train a liveness detection model tailored to the actual data used to supplement existing algorithms. If we continue to expand the latest attack images into the CLASS-8 dataset, we believe that in practical applications in the financial sector, CLIPC8 can complement commercial privatized algorithms, improving the detection rate of abnormal images in specific scenarios.The main contributions of this paper are as follows:
\\\indent$\diamond$  For the first time, to apply the concept of image-text pairing and contrastive learning to the task of face liveness detection. Through experimental analysis, we have found that integrating categorical text descriptions into image analysis significantly improves the accuracy of live detection. Test results on five public datasets confirm that this method can effectively identify live attack images, and has an extremely low false recognition rate for real faces.
\\\indent$\diamond$  We conducted an analysis of the detection rates of commercial live detection algorithms from four companies using the CLASS-8 dataset. The study found that these algorithms cannot effectively recognize live attack images in certain scenarios. This indicates that in practical applications, encountering attack samples not in the training dataset might lead to misrecognition.
\\\indent$\diamond$  This research also analyzed the commercial live detection algorithms of four companies using five commonly used live detection datasets: NUAA, CASIA-FASD, Replay-Attack, OULU-NPU, and MSU-MFSD, and compared the detection rates of each algorithm.

This paper suggests that when applying face liveness detection technology in the financial sector, in addition to collaborating with commercial companies to use their liveness detection algorithms, it is also necessary to collect and establish private datasets specific to the field for training. This approach can help avoid the problem of low detection rates in certain specific scenarios when relying on commercial algorithms.

\Section{Related Work}\label{sec:related}
Previous studies~\cite{08liao2023domain,09chaudhry2023transfas,10young2023can,11wang2022face,12qiao2022fgdnet,13liu2023fm,14liu2023ma,15samar2022multi,16george2021effectiveness,17yu2021transrppg} have delved into the realm of image liveness detection, initiating their explorations with the Transformer model~\cite{18vaswani2017attention}. Among these, Samar \etal \cite{15samar2022multi} harnessed the intrinsic RGB and IR depth information in images, utilizing a dual Transformer model architecture to discern liveness attack patterns. Wang \etal \cite{11wang2022face} adopted Transformer encoder layers as the core network, integrating Cross-layer Relation-aware Attentions (CRA) and hierarchical feature fusion techniques to augment liveness detection efficacy. George \etal \cite{16george2021effectiveness} observed a tendency for overfitting to training datasets in most methods, leading to diminished detection performance in novel attack scenarios and environments. Their study focused on zero-shot liveness detection tasks, leveraging the transfer learning potential of the Vision Transformer model. Young \etal \cite{10young2023can} capitalized on Swin Transformer layers and adaptive windows to capture nuanced facial structure representations, effectively employing both global attention mechanisms and local detail features. Liu \etal \cite{13liu2023fm} introduced the FM-ViT (Flexible Modal Vision Transformer) algorithm, enhancing face liveness detection systems' performance through multimodal training data utilization. Liao \etal \cite{08liao2023domain} highlighted the challenge of addressing attack samples from unknown domains in liveness detection, proposing the Domain-invariant Vision Transformer (DiVT) that employs efficient Vision Transformer (ViT) based modules for extracting comprehensive global and local distribution patterns of liveness attacks. Yu \etal \cite{17yu2021transrppg} devised a dual-branch Transformer model for distinct feature extraction from facial and background regions in images, followed by a Transformer layer for binary classification prediction between real individuals and 3D mask attacks. Wang \etal \cite{19wang2022learning} observed that real and fake faces are highly similar in facial appearance, while temporal features between continuous frames in videos can help differentiate between the real and fake, using ViT as the backbone network to learn time-based features for distinguishing subtle clues between the two. Qiao \etal \cite{12qiao2022fgdnet} argued that most liveness detection algorithms focus on binary classification and typically fail to accomplish fine-grained classification, employing a Transformer network structure for feature extraction and attention modules to enhance focus on distant information. Chaudhry \etal \cite{09chaudhry2023transfas} introduced the Video Vision Transformer (VVT) structure to extract feature values from a set of video frames, which, after adding positional information, are processed through a Transformer layer, achieving lower error rates on multiple datasets.\\
\indent These methodologies primarily consider liveness detection from a visual perspective, overlooking the potential contributions of textual information in augmenting image classification accuracy. OpenAI's CLIP (Contrastive Language-Image Pre-Training) model~\cite{20radford2021learning}, achieves zero-shot transfer in downstream tasks by learning to predict a predefined set of object categories, using a dataset comprising 400 million image-text pairs sourced from the internet. The CLIP model's test outcomes demonstrate that, even without the 1.28 million training samples from ImageNet, it achieves an accuracy comparable to the original ResNet-50 model trained on the same dataset. Yu \etal \cite{21yu2022coca} introduced the CoCa (Contrastive Captioners) model, which utilizes large-scale image-text pairs and contrastive learning. This model discerns the connections between images and corresponding texts from positive pairs and distinguishes them using negative pairs. Through iterative training, the model extracts salient visual and semantic features from an extensive dataset of images and texts. Post-encoder fine-tuning, the model reached a top-1 accuracy of 91.0\% on ImageNet, setting a new benchmark. The remarkable performance of image-text pre-training models such as CLIP and CoCa on the ImageNet dataset underscores the value of textual information in enhancing image classification tasks. Building on this insight, this paper integrates the principle of contrastive learning into the liveness detection domain, fine-tuning the CLIP model on the newly proposed CLASS-8 dataset.

\begin{figure}[htp]
	\centering
	\begin{subfigure}{0.22\linewidth}
		\includegraphics[width=\linewidth]{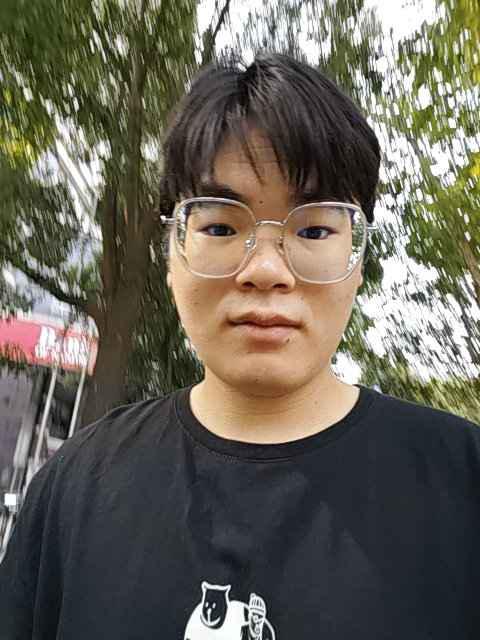}
		\caption{}
		\label{figure_class8_dataset_a}
	\end{subfigure}
	\hspace{2pt} 
	\begin{subfigure}{0.22\linewidth}
		\includegraphics[width=\linewidth]{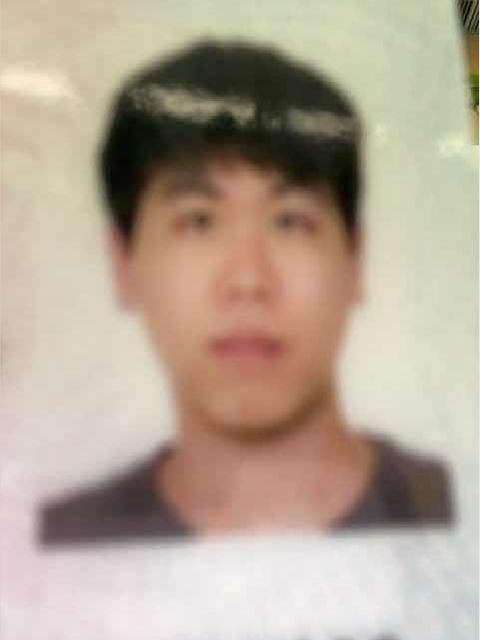}
		\caption{}
		\label{figure_class8_dataset_b}
	\end{subfigure}
	\hspace{2pt}
	\begin{subfigure}{0.22\linewidth}
		\includegraphics[width=\linewidth]{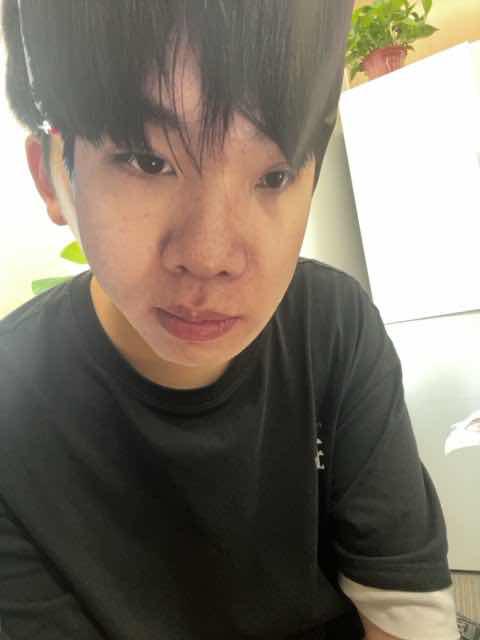}
		\caption{}
		\label{figure_class8_dataset_c}
	\end{subfigure}
	\hspace{2pt}
	\begin{subfigure}{0.22\linewidth}
		\includegraphics[width=\linewidth]{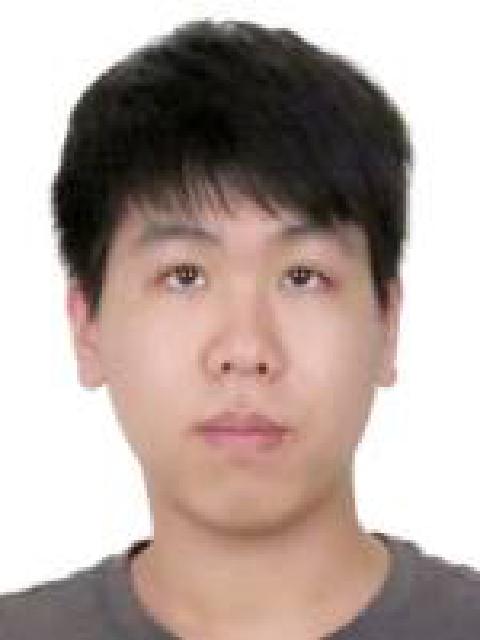}
		\caption{}
		\label{figure_class8_dataset_d}
	\end{subfigure}
	
	\begin{subfigure}{0.22\linewidth}
		\includegraphics[width=\linewidth]{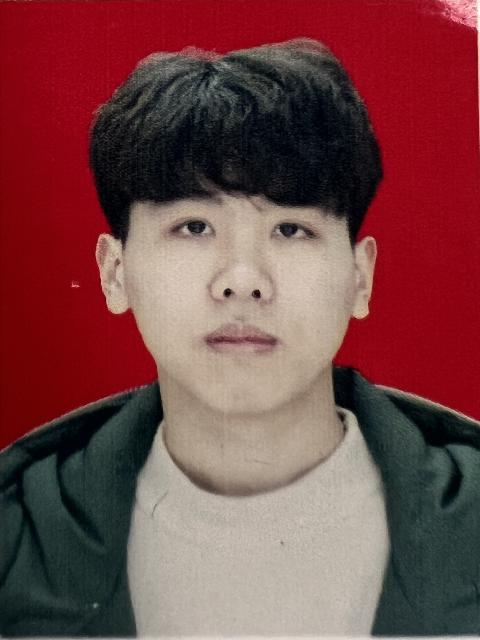}
		\caption{}
		\label{figure_class8_dataset_e}
	\end{subfigure}
	\hspace{2pt}
	\begin{subfigure}{0.22\linewidth}
		\includegraphics[width=\linewidth]{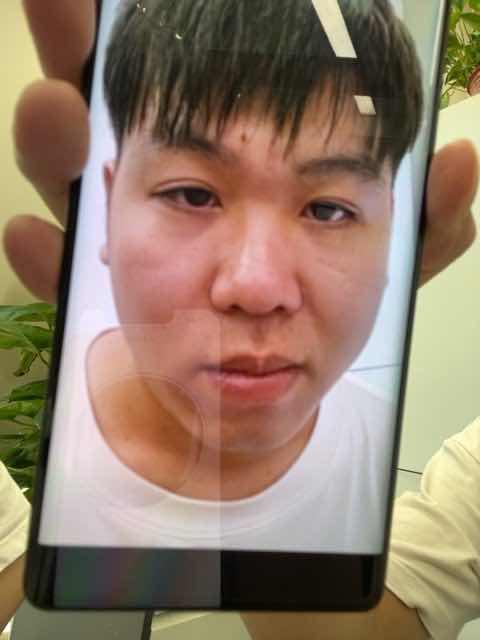}
		\caption{}
		\label{figure_class8_dataset_f}
	\end{subfigure}
	\hspace{2pt}
	\begin{subfigure}{0.22\linewidth}
		\includegraphics[width=\linewidth]{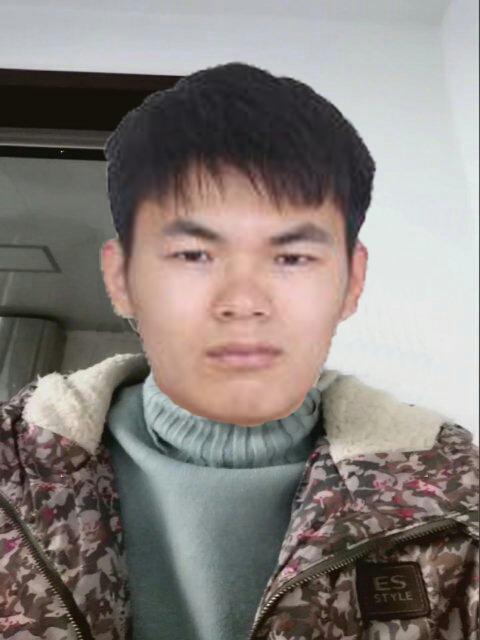}
		\caption{}
		\label{figure_class8_dataset_g}
	\end{subfigure}
	\hspace{2pt}
	\begin{subfigure}{0.22\linewidth}
		\includegraphics[width=\linewidth]{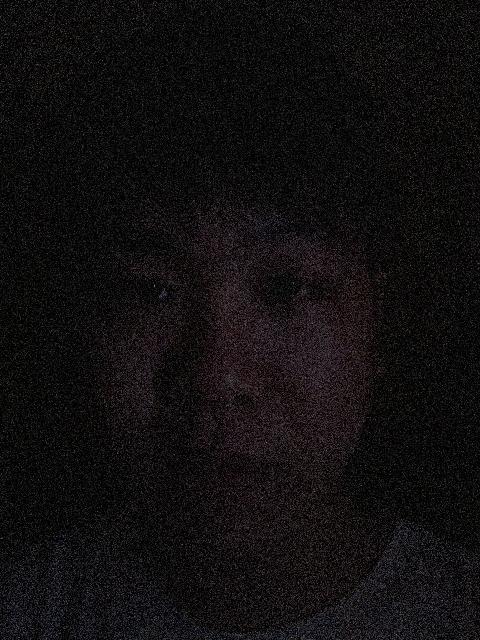}
		\caption{}
		\label{figure_class8_dataset_h}
	\end{subfigure}
	\raggedright
	\footnotesize Note: The above 8 samples were photographed and produced by the authors of this paper and are used only for demonstrating data examples in the paper.
	\caption{CLASS-8 dataset}
	\label{figure_class8_dataset}
\end{figure}

\Section{Introduction to the CLASS-8 Dataset}\label{sec:dataset}
\SubSection{Image Dataset} \label{sec_image_dataset}
General face recognition algorithms are usually developed through training and performance evaluation on various open datasets. However, in practical applications, it's essential to evaluate these algorithms' ability to process images taken with different devices, under different lighting conditions, and from various angles. Specifically addressing the issue of commercial liveness detection algorithms in the financial sector failing to effectively recognize attack images and behaviors in certain scenarios, this paper, drawing from practical experience in the financial field, proposes the CLASS-8 dataset.We believe that real liveness detection should emphasize immediacy and authenticity. Immediacy means that pre-captured photos, especially those obtained through the black industry chain (like online verification photos, ID photos, and document photos), should not be used as real data. Authenticity refers to the spontaneity in the face recognition process, where using facial data of others, such as through print attacks or screen recapturing, is not permissible. This dataset aims to distinguish between compliant and suspicious behaviors to safeguard the security of face recognition processes in the financial sector, summarizing 8 types of images. Fig.~\ref{figure_class8_dataset} presents typical samples of these 8 categories:~\ref{figure_class8_dataset}\subref{figure_class8_dataset_a}  normal face photos taken with cameras of devices like smartphones or tablets,~\ref{figure_class8_dataset}\subref{figure_class8_dataset_b} photos displayed in front of a camera by holding an identify card
photo,~\ref{figure_class8_dataset}\subref{figure_class8_dataset_c} photos with face printed on paper or 3D masks,~\ref{figure_class8_dataset}\subref{figure_class8_dataset_d} online verification photos with white backgrounds,~\ref{figure_class8_dataset}\subref{figure_class8_dataset_e} personal ID photo mostly with red or blue backgrounds,~\ref{figure_class8_dataset}\subref{figure_class8_dataset_f} recaptured photos displayed through screens of smartphones or computers,~\ref{figure_class8_dataset}\subref{figure_class8_dataset_g} photos altered using software like Photoshop, and~\ref{figure_class8_dataset}\subref{figure_class8_dataset_h} face recognition photos taken in dark environments. Among these, mask wearing and screen recapturing are common attack methods, while the others represent less common approaches.\\
\indent In the process of dataset selection and categorization, multiple aspects were comprehensively considered. The financial sector often involves the use of personal ID cards in business transactions, and these scenes, if not properly secured, can be illegally used by others. These ID photos can be easily obtained through HR systems, social media profile pictures, etc. Additionally, online verification photos can be acquired through various channels. In simulated liveness attack scenarios, most attacks involve individuals statically displaying their personal ID or document photos in front of the front camera of a phone, deceiving the face recognition system. Represented by categories~\ref{figure_class8_dataset}\subref{figure_class8_dataset_b},~\ref{figure_class8_dataset}\subref{figure_class8_dataset_d}, and~\ref{figure_class8_dataset}\subref{figure_class8_dataset_e}. Screen recapturing is also a common attack method, characterized by features like moiré patterns and the borders of the phone or computer screen, is categorized under~\ref{figure_class8_dataset}\subref{figure_class8_dataset_f}. Photos presented before cameras through printing or mask wearing are included in category~\ref{figure_class8_dataset}\subref{figure_class8_dataset_c}.\\
\indent Most commonly used face recognition datasets collect training and testing samples in well-lit environments. However, this paper finds that in low-light environments, the success rate of static or dynamic face images using screen recapturing in attacking liveness detection systems is higher than in well-lit situations, as represented in category~\ref{figure_class8_dataset}\subref{figure_class8_dataset_h}. Category~\ref{figure_class8_dataset}\subref{figure_class8_dataset_g} demonstrates Photoshopped images, commonly used in injection-style attacks.
The establishment of normal facial data~\ref{figure_class8_dataset}\subref{figure_class8_dataset_a} takes into account the specific characteristics of facial recognition systems used in the financial sector, focusing only on clear images with uniform data formats and resolutions. The CLASS-8 dataset encompasses both common and uncommon attack methods, aligning with the actual use of facial recognition systems in the financial sector.
\SubSection{Textual Description Information for Image Classification} \label{sec_textual_des}
This paper employs English text to describe the above-mentioned eight categories of images, endeavoring to broadly summarize the common characteristics of each image category. The abbreviations for each category are as follows: Tab.~\ref{figure_class8_dataset}\subref{figure_class8_dataset_a} Normal Facial Photo;~\ref{figure_class8_dataset}\subref{figure_class8_dataset_b} Identify Card Photo;~\ref{figure_class8_dataset}\subref{figure_class8_dataset_c} Face Mask;~\ref{figure_class8_dataset}\subref{figure_class8_dataset_d} Online Verification Photo~\ref{figure_class8_dataset}\subref{figure_class8_dataset_e} Personal ID Photo;~\ref{figure_class8_dataset}\subref{figure_class8_dataset_f} Recaptured Photo;~\ref{figure_class8_dataset}\subref{figure_class8_dataset_g} Photoshopped Image;~\ref{figure_class8_dataset}\subref{figure_class8_dataset_h} Dark Environment. For detailed textual description information and corresponding relationships, refer to Tab.~\ref{des_text_dataset}.
\begin{table}[t!]
\centering
\normalsize
\resizebox{0.5\textwidth}{!}{
\begin{tabular}{c|m{0.4\textwidth}}
\hline
Category & Descriptive Text for Each Category\\\hline
(a) & A normal selfie photo of a person taken with the front camera of a mobile phone under well-lit conditions, and the image has not been altered.\\\hline
(b) & Partial or complete photos of identification cards, with visible address and identification number information in the photos.\\\hline
(c) & A face photo printed on paper, a person wearing a face mask, and partial information of the hands may be visible in the photo.\\\hline
(d) & Online verification of citizens photo with a white background, the person in the photo is not wearing a hat and has black hair.\\\hline
(e) & A person's work identification photo, usually with a person in a suit and a white, blue or red background.\\\hline
(f) & Displaying personal photos or identification card photos on a mobile phone, where the attack samples may reveal the borders of the phone or computer, moiré patterns, and partial information of the hands.\\\hline
(g) & The photo has been manipulated using PS, resulting in distorted details in certain areas.\\\hline
(h) & The photo was taken in a dark environment, making it impossible to discern facial details. The entire frame is dominated by black, providing no visible information.\\\hline
\end{tabular}
}
\vspace{-2mm}
\caption{Descriptive Text for CLASS-8 dataset.  \vspace{-3mm}}			
\label{des_text_dataset}
\end{table}

\begin{figure}[htbp]
	\centering
	\includegraphics[width=0.5\textwidth]{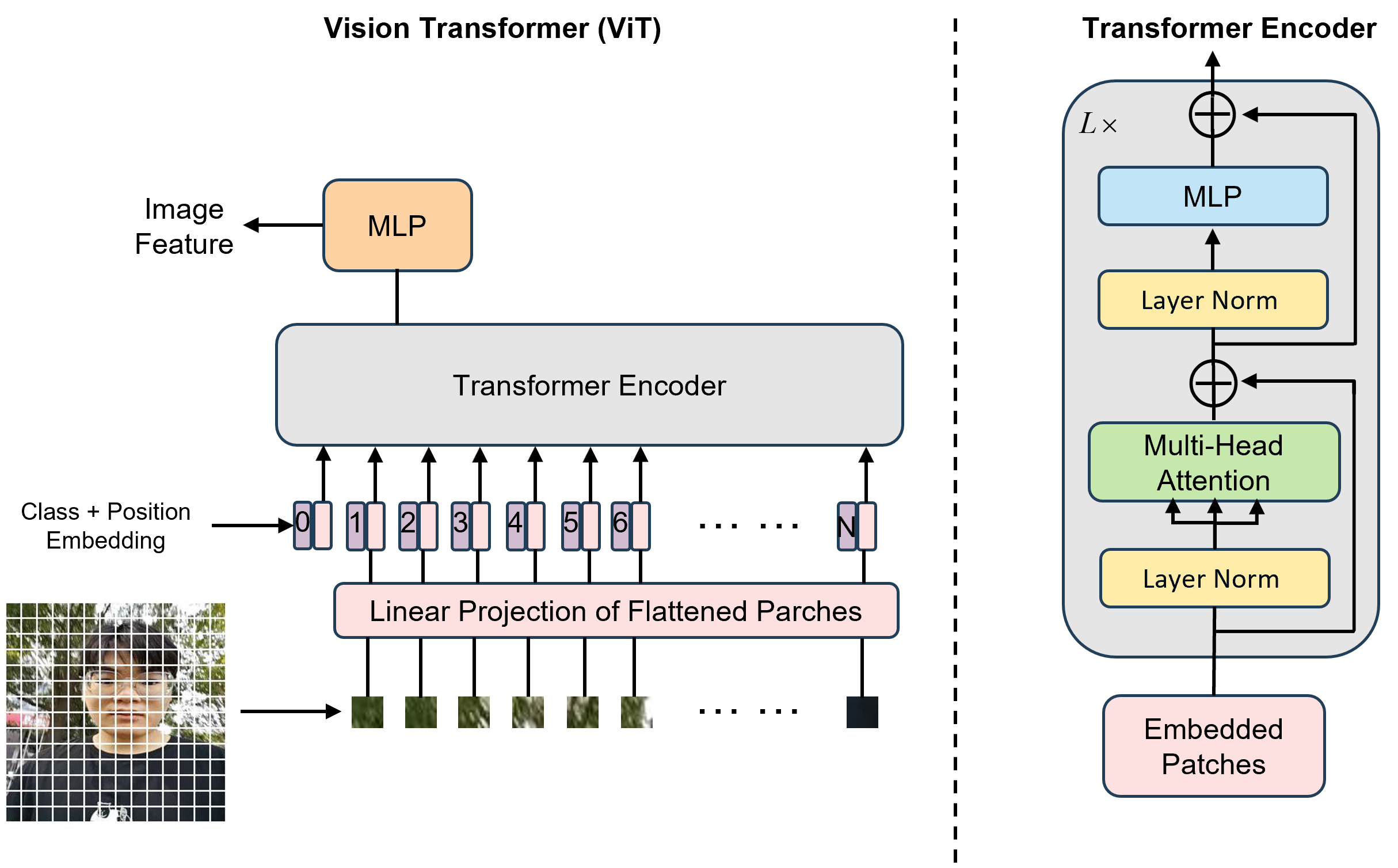} 
	\caption{ViT architecture}
	\label{figure_vit_model} 
\end{figure}

This paper aims to combine the CLASS-8 dataset with image-text contrastive learning, covering a variety of uncommon attack behaviors, while utilizing textual information to improve the detection rate of the live detection algorithm.
\Section{CLIPC8}
\label{sec:method}
\SubSection{Image Encoder} \label{sec_image_encoder}
In this paper, we employ the Vision Transformer (ViT) as the image encoder. Fig.~\ref{figure_vit_model} illustrates the structure and processing flow of ViT. Initially, an image of arbitrary resolution \((h, w)\) is inputted and then scaled to a fixed resolution \((M,M)\). Subsequently, based on predefined model parameters, the image is segmented into multiple small pieces with a resolution of \((c,c)\). These segments, derived from the original image, are referred to as \( X_i \), and their quantity \( N \) can be calculated using:
\begin{equation}
	N = \frac{M \times M}{c^2}
\end{equation}

Following this, we use a trainable linear projection layer \(E\) to map these \(N\) pieces of \( X_i \), where $i \leq N$
, to an embedding layer. To enable the model to learn the differences and commonalities between various categories, we incorporate classification information \( X_{class} \) into the image encoding. Moreover, to maintain the positional information of the image, it is necessary to add positional labels \( P_{pos} \):
\begin{equation}
	P_E = [X_{\text{class}}, X_E, X_1, X_2, \ldots, X_N] + P_{\text{pos}}
\end{equation}

A Transformer module is composed of a Multi-head Self-Attention layer (MSA) and a Multi-Layer Perceptron (MLP). We employ layer normalization (LN) to normalize the input data across the feature dimension. The output of \( l_{th} \) layer can be expressed in the following manner:
\begin{equation}
	y^l = \text{MSA}(\text{LN}(P^l)) + P^l
\end{equation}
\begin{equation}
	P^{(l+1)} = \text{MLP}(\text{LN}(y^l)) + y^l
\end{equation}

Here, \( P^l \) denotes the input to the \( l_{th} \) layer encoder, and \(y^l\) represents the intermediate variable.

\SubSection{The CLIPC8 Method} \label{sec_clipc8_method}
In this paper, we introduce a liveness detection method named CLIPC8, which is trained based on image-text pairs. We utilized the pre-trained parameters of the CLIP ViT-L/14 model and conducted training on the CLASS-8 dataset to leverage supervisory signals contained in natural language for training an image classification model. The features obtained from our training include not only visual characteristics but also multimodal features, which facilitate transfer learning across different datasets. Images are processed through the image encoding layer constructed by ViT to generate image feature \(I\), while texts are processed through the text encoding layer built by Transformer to produce text feature \(T\). We calculate the cosine similarity \(S\) between the text features and image features, and use the softmax function to process this similarity  \(S\), converting the network's output into probabilities \(C\) for each category, where the highest scoring category is considered the current category. In the 8 categories proposed in Sec.~\ref{sec_image_dataset},  Fig.~\ref{figure_class8_dataset_a} \enquote{Normal Facial Photo} is classified as \enquote{Real}, while the other 7 categories are considered as liveness attacks. The categorization process can be represented by the following formula: 

\begin{equation}
	C = \text{softmax}(\text{cosine\_similarity}(I, T))
\end{equation}
\begin{equation}
	S = \text{cosine\_similarity}(I, T) = \frac{I \cdot T}{\| I \|_2 \cdot \| T \|_2}
\end{equation}
\begin{equation}
	C_i = \text{softmax}(S_i) = \frac{e^{S_i}}{\sum_{j=1}^{8} e^{S_j}}
\end{equation}
\begin{equation}
	L = \text{MAX}(C_i) = \left\{ \begin{array}{ll}
		Real, & \text{if}\ i=\text{index}(Real) \\
		Fake, & \text{if}\ i\neq\text{index}(Real)
	\end{array} \right.
\end{equation}

Here \(L\) represents the final category. In our experiments, $1 \leq i \leq 8$ corresponds to the number of categories in the CLASS-8 dataset, and \(C_i\) represents the probability of each category. If the \(i\) with the highest score equals the sequence number of the real person's label in the sample, then the algorithm identifies it as a real person; otherwise, it is identified as a liveness attack.

\begin{figure}[htbp]
	\centering
	\includegraphics[width=0.5\textwidth]{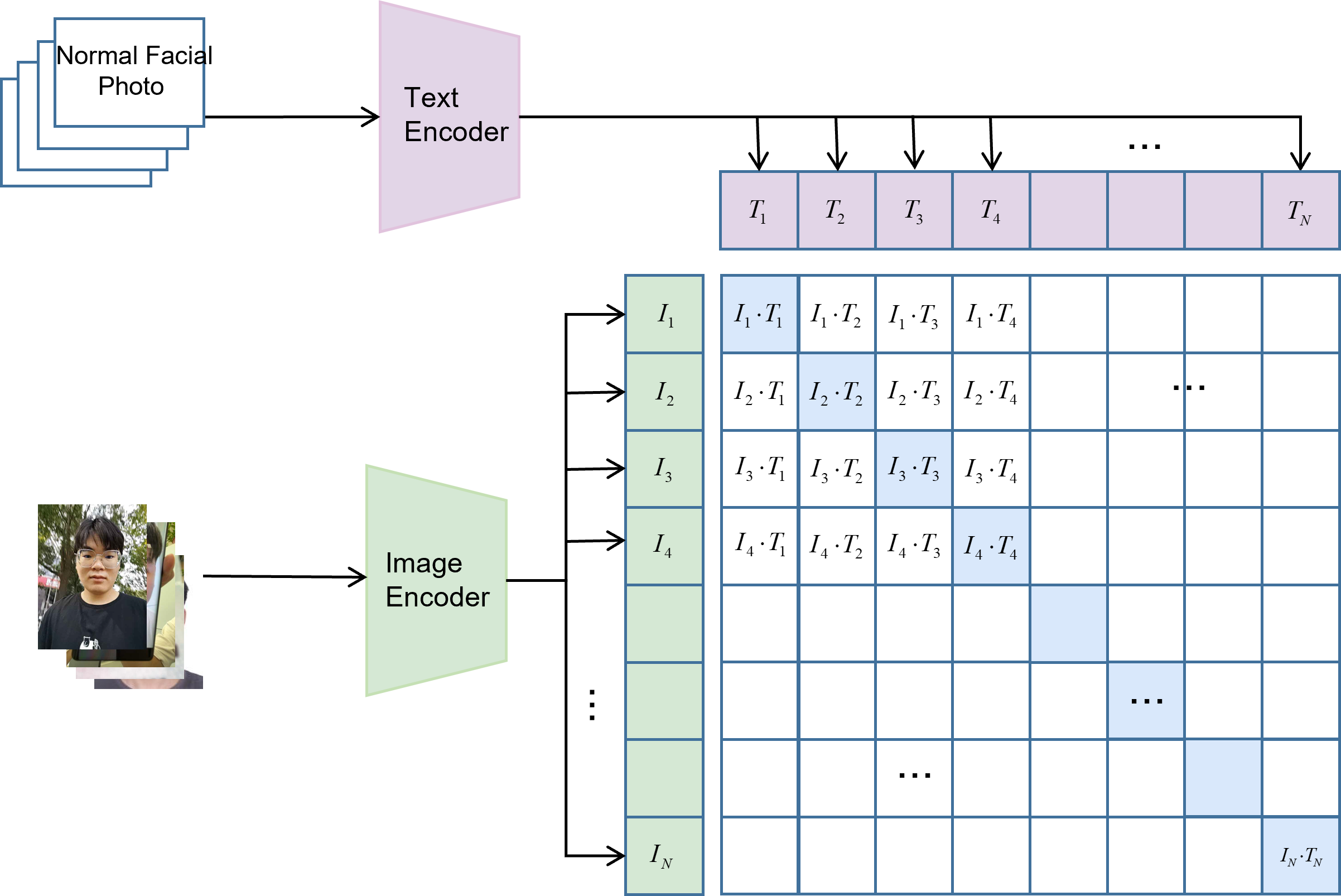} 
	\caption{Image-Text Contrastive Learning Based on the CLASS-8 Dataset}
	\label{figure_Contrastive_Learning} 
\end{figure}

Fig.~\ref{figure_Contrastive_Learning} illustrates the training process of the CLIPC8 algorithm. During the training phase, the concept of contrastive learning is utilized to compare positive samples (images and corresponding classification texts) with negative samples (images and texts of different classifications). The ViT structure is employed as the backbone network. The model processes image-text pairs as inputs and computes their similarity in the vector space. The CrossEntropyLoss function encourages higher scores for positive samples and lower scores for negative samples. Through contrastive learning, the CLIPC8 model learns a shared representation linking images and texts. 

\begin{figure}[htbp]
	\centering
	\includegraphics[width=0.5\textwidth]{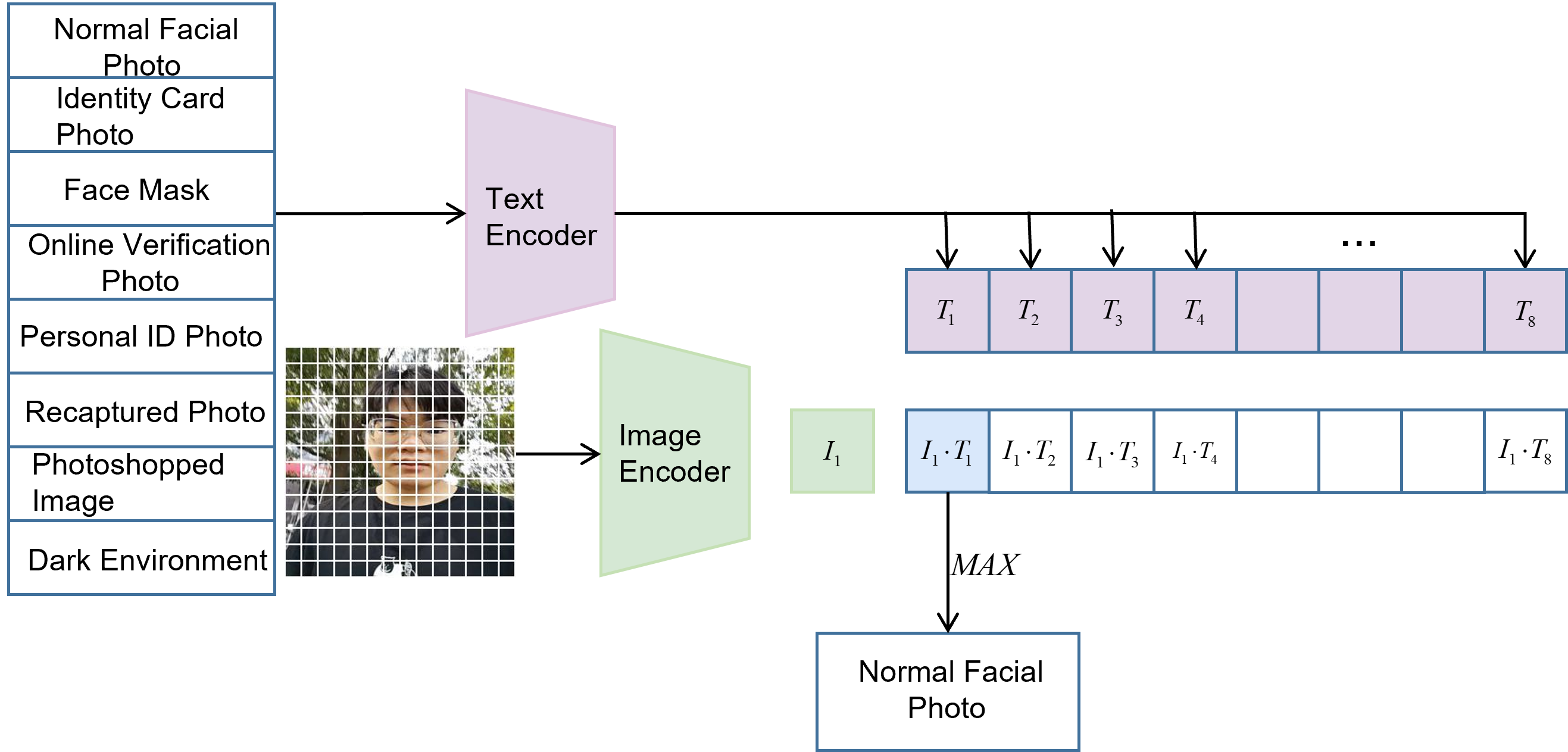} 
	\caption{Face liveness detection based on CLIPC8}
	\label{figure_liveness_detection} 
\end{figure}

Fig.~\ref{figure_liveness_detection} shows the input and output of the model during the prediction phase. To determine whether the subject is a real human or a live attack, the CLIPC8 model receives an image along with eight sets of text classification prompts. The model calculates the similarity between the input image and each classification text prompt, and the category corresponding to the highest similarity is the predicted category.


\Section{Experiments}
\label{sec:experiments}
\SubSection{Experimental Environment} \label{sec_ex_en}
The experimental operating system is Centos 7.6, with the integrated development environment being Python 3.9.7, utilizing the Pytorch deep learning framework. The hardware configuration comprises an Intel(R) Xeon(R) Gold 6326 CPU @ 2.90GHz CPU and four NVIDIA Tesla T4 GPUs. Tab.~\ref{tab_Running_environment} presents the operating environments of four different algorithms, where configuration discrepancies do not affect the returned results of the algorithms.
\begin{table}[t!]
	\centering
	\normalsize
	\resizebox{0.5\textwidth}{!}{
		\begin{tabular}{c|c|c|c|c}
			\hline
			Company & Algorithm Version & Operating System & CPU Cores & Memory/GB \\\hline
			SenseTime & 2.2.1 & Centos7.6 & 8 & 32 \\
			SenseTime & 2.2.5 & Centos7.6 & 12 & 32 \\
			Megvii & 3.1.98 & Centos7.6 & 4 & 32 \\
			Megvii & 3.2.23 & Centos7.6 & 12 & 32 \\
			Tencent & 1.0.162 & Centos7.6 & 16 & 32 \\
			CloudWalk & 1.1.4 & Centos7.6 & 8 & 32 \\\hline
		\end{tabular}
	}
	\vspace{-2mm}
	\caption{Running environment and configuration of the four commercial liveness detection algorithms.\vspace{-3mm}}
	\label{tab_Running_environment}
\end{table}
\SubSection{Experimental Details} \label{sec_de_ex}
The ViT-L/14 model serves as the foundational model for the CLIPC8 algorithm, with the text encoder consisting of a Transformer model with 12 layers, each containing 12 multi-head self-attention modules and 3072 hidden layer nodes. The image encoder employs a ViT module with 24 layers, 16 multi-head self-attention modules, and 4096 hidden layer nodes. A total of 100 training iterations were conducted, with a batch size set to 16. In image processing, considering the live detection model's learning of associations between images and text pairs, data augmentation methods like random cropping and color jittering that alter the intrinsic live detection judgment of images were avoided. Only random horizontal flipping and random erasure were used to introduce variability and diversity in the images. During training, the text encoder's parameters were frozen, and only the image encoder was trained using the Adam optimizer~\cite{22kingma2014adam}. The initial learning rate and weight decay were set at 4e-6 and 0.05, respectively, with the learning rate varying according to the CosineAnnealingLR~\cite{23loshchilov2016sgdr} method. The learning rate decay cycle was set for every 10 rounds, with the minimum learning rate set to 0.01 of the initial rate. This method adjusts the learning rate in a cosine annealing manner each cycle, thereby aiding the model in more effective convergence and generalization.
\SubSection{Test Dataset} \label{sec_di_td}
The NUAA dataset was collected via computer cameras and includes frontal facial postures of 15 subjects in various environments. The dataset was compiled using camera paper and A4 paper to print counterfeit faces, which were then recaptured with computer cameras. The CASIA-FASD dataset contains 600 videos recorded by 50 individuals under different lighting and scene conditions, comprising 12 videos each. The primary attack methods include printed photo attacks, tablet computer video playback, and remove the eyes from the printed photo of the face. The OULU-NPU dataset consists of 5940 real and spoofed videos, recorded using the front cameras of 6 smartphones under three different lighting conditions and background scenes. The Replay-Attack dataset comprises 1200 videos, photos of real people and live attacks involving 50 participants under different lighting scenarios, including real faces, printed photo replay, and mobile and tablet computer video replay. The MSU-MFSD dataset includes 35 participants, each recording 2 real videos and 6 spoof videos, totaling 280 videos, covering printed photo attacks and screen video playback attacks.
\SubSection{Evaluation Method} \label{sec_eval_method}
The detection rate \(U\) is used as the metric to assess the performance of the live detection algorithm on public datasets and the CLASS-8 dataset. The formula for calculation is as follows: 
\begin{equation}
U = \frac{R}{Q}
\end{equation}
\indent Regarding the positive samples in the dataset, \(R\) denotes the number of samples that the algorithm classifies as pass. For negative samples, \(R\) represents the number of samples that the algorithm classifies as reject, while \(Q\) indicates the total number of test samples. Therefore, an increase in the \(U\) value for both positive and negative samples indicates an enhancement in the detection capability of the algorithm.
\SubSection{Preserving Background Information in Facial Images} \label{sec_preserve_face}
In the process of data handling, previous studies ~\cite{08liao2023domain,24wang2022robustness,25ren2023face} first crop and align faces before using this processed data for algorithm training. Although this process enhances focus on local facial information, it overlooks auxiliary information outside of the face. Yang \etal \cite{01yang2019face} note that liveness detection algorithms need to identify subtle evidences (such as edges, mesh patterns, reflective artifacts, etc.) to differentiate between real faces and liveness attack. This section primarily focuses on both local facial information and overall image information, aligning and cropping the facial regions in the NUAA dataset to form the NUAA-CutFace dataset. It then compares and analyzes the detection rates of four algorithms and CLIPC8 on both the NUAA and NUAA-CutFace datasets. Tab.~\ref{tab_exper_on_nuaa} displays the detection rates of these algorithms, calculated using the method described in Sec.~\ref{sec_eval_method}, where \enquote{Real} represents the positive samples in the dataset, and \enquote{Fake} represents the negative samples. Zero-shot CLIPC8 indicates that the algorithm is trained only on the CLASS-8 dataset and has not been exposed to any data from the NUAA dataset. From the analysis in Tab.~\ref{tab_exper_on_nuaa}, we find that the average detection rate of each algorithm on the NUAA dataset is higher than that on the NUAA-CutFace dataset. Therefore, this paper posits that the introduction of global information helps the algorithms more accurately distinguish between real faces and liveness attacks. All subsequent experimental results are based on images with retained background information.
\begin{table*}[t!]
	\centering
	\normalsize
	\resizebox{\textwidth}{!}{
		\begin{tabular}{c|c|c|c|c|c|c|c|c}
			\hline
			DataSets & Label & SenseTime$^{1}$ & SenseTime$^{2}$ & Megvii$^{1}$ & Megvii$^{2}$ & Tencent & CloudWalk & Zero-shot CLIPC8 \\\hline
			\multirow{2}{*}{NUAA-Cutface} & Real &88.98&	83.65&	74.48&	52.17&	41.91&	99.28&	62.76\\
			                             & Fake & 97.82&	98.82&	99.64&	99.91&	99.49&	3.34&	99.84\\\hline
			\multirow{2}{*}{NUAA}        & Real & 87.33&	99.94&	88.01&	79.33&	90.09&	100.00&	95.87\\
									   & Fake & 99.99	&99.99&	99.97&	99.95&	99.95&	0.00&	91.49\\\hline
		\end{tabular}
	}
	\raggedright
	\footnotesize Note: Sensetime$^{1}$ represents the algorithm version 2.2.5, Sensetime$^{2}$ represents the algorithm version 2.2.1, Megvii$^{1}$ represents the algorithm version 3.2.23, and Megvii$^{2}$ represents the algorithm version 3.1.98.
	\vspace{-2mm}
	\caption{Running environment and configuration of the four commercial liveness detection algorithms.\vspace{-3mm}}
	\label{tab_exper_on_nuaa}
\end{table*}
\SubSection{Detection Rate Evaluation Based on CLASS-8 Dataset} \label{sec_detection_rate}
This section primarily introduces the detection rates of four commercial liveness detection algorithms on the CLASS-8 dataset. As seen in Tab.~\ref{tab_detectionrate_onclass8}, CloudWalk  scores the highest in detecting real person photos, but the lowest in detecting attack samples. A preliminary assessment suggests that CloudWalk's proprietary liveness detection algorithm focuses mainly on real person recognition, while neglecting various types of liveness attacks. Aside from CloudWalk, the other three companies exhibit high detection rates for identity card replication and recaptured screen attacks, but generally lower rates for masks attack, Photoshopped image, dark environments, and other uncommon scenarios. Additionally, SenseTime scores lower in the categories of online verification photo and personal ID photo. Considering the performance of the four companies on the CLASS-8 dataset, Megvii slightly outperforms the others.
\indent The CLASS-8 dataset comprehensively considers various types of attack images encountered in the actual face recognition process, including uncommon liveness attack image data such as online verification photo, identify card photo, Photoshopped image, and dark environments. The five public datasets described in Sec.~\ref{sec_di_td} do not cover such data. The proposed CLASS-8 dataset enriches the range of liveness attack behaviors, providing new insights for the future establishment of face liveness attack datasets.
\begin{table*}[t!]
	\centering
	\begin{tabular}{c|c|c|c|c|c|c}
		\hline
		Classes & SenseTime$^{1}$ & SenseTime$^{2}$ & Megvii$^{1}$ & Megvii$^{2}$ & Tencent & CloudWalk\\\hline
		(a)& 99.53&	99.88&	98.77&	99.65&	94.65&	99.98\\
		(b)&99.50&	99.91&	99.60&	99.75&	99.23&	32.18\\
		(c)&87.79&	33.72&	94.19&	87.79&	88.95&	1.16\\
		(d)&27.10&	93.21&	98.24&	89.85&	99.47&	0.61\\
		(e)&33.14&	41.14&	98.49&	98.15&	95.71&	0.00\\
		(f)&99.67&	99.86&	99.21&	99.34&	99.21&	5.03\\
		(g)&66.59&	67.30&	86.03&	76.51&	78.10&	7.03\\
		(h)&92.28&	64.81&	71.08&	71.39&	67.79&	52.13\\\hline
	\end{tabular}
	
	\raggedright
	\footnotesize Note: Sensetime$^{1}$ represents the algorithm version 2.2.5, Sensetime$^{2}$ represents the algorithm version 2.2.1, Megvii$^{1}$ represents the algorithm version 3.2.23, and Megvii$^{2}$ represents the algorithm version 3.1.98.(a) Normal Facial Photographs;(b) Partial or Complete Identity Card Photographs;(c) Printed Facial Photographs with Masks;(c) Printed Facial Photographs with Masks;(d) Online Verification Photographs against a White Background;(e) Photographs with White, Blue, or Red Backgrounds;(f) Personal or Identity Card Photographs Displayed on Mobile Devices;(g) Photoshopped Images with Distorted Details;(h)Photographs in Dark Environments.
	\vspace{-2mm}
	\caption{Detection rates of commercial liveness detection algorithms on the CLASS-8 dataset.\vspace{-3mm}}
	\label{tab_detectionrate_onclass8}
\end{table*}
\SubSection{Assessment of Detection Rates Based on Five Public Datasets} \label{sec_assessment_dec}
This section primarily presents the detection rate evaluation results of various commercial liveness detection algorithms across five public datasets, with results illustrated in Tab.~\ref{table_comparison_public}.

\indent Analyzing these datasets, except for CloudWalk, the other three companies have achieved high detection rates for liveness attack images in the NUAA dataset and also perform well in recognizing real human images. The OULU-NPU dataset contains attack samples that are difficult to distinguish by the naked eye, resulting in lower detection rates for these samples across the companies. Tencent's algorithm has a lower detection rate for real human samples in the Replay-Attack dataset, but other companies have achieved a better balance between real human and attack samples. In the CASIA-FASD dataset, SenseTime performs excellently, but other companies have lower detection rates for real human samples. For the MSU-MFSD dataset, none of the companies have reached the ideal level of detection rates for both real humans and liveness attack samples. Overall, each company's liveness detection algorithms have their strengths and weaknesses across the five datasets, with none being absolutely dominant. This table's result shows that the liveness detection algorithms of SenseTime and Megvii perform better than those of Tencent and CloudWalk across the five datasets. CloudWalk's poor detection rate for attack images may indicate that the company focuses more on implementing liveness detection at the facial recognition SDK level, paying less attention to research on private deployment liveness detection algorithms.
\indent As mentioned in Sec.~\ref{sec_preserve_face}, the Zero-shot CLIPC8 algorithm is trained on the CLASS-8 dataset, which does not contain any images from the five public datasets. Our proposed CLIPC8 algorithm performs well in most datasets, especially in terms of the positive sample recognition rate for real human images. This suggests that the CLIPC8 algorithm can learn to distinguish between liveness attacks and real human images from image-text pairs, effectively recognizing the common features of real human images. Since the CLASS-8 dataset focuses on the significant differences between different attack methods, and the OULU-NPU dataset's liveness attack images have almost no significant attack features (such as screen edges, moiré patterns, abnormal behavior) other than brightness differences, the detection rate of CLIPC8 on this dataset is lower.
After integrating the training sets of the five datasets into CLASS-8, we finetuned the CLIPC8 algorithm and verified it on the test sets of these five datasets. As shown in the \enquote{Trained CLIPC8} column of Tab. \ref{table_comparison_public}, the algorithm achieved 100\% detection rates for both real human and liveness attack images in the NUAA and MSU-MFSD test datasets. Comparing the \enquote{Trained CLIPC8} with the commercial liveness detection algorithms of the other four companies, CLIPC8 leads in detection rates for real humans and attacks across all five datasets, significantly outperforming the other four companies.
\begin{table*}[t!]
	\centering
	\begin{tabular}{cccccccccc}
		\hline
		DataSets & Label & SenseTime$^{1}$ & SenseTime$^{2}$ & Megvii$^{1}$ & Megvii$^{2}$ & Tencent & CloudWalk & \begin{tabular}[c]{@{}c@{}}Zero-shot \\CLIPC8\end{tabular} & \begin{tabular}[c]{@{}c@{}}Trained \\CLIPC8\end{tabular}\\\hline
		\multirow{2}{*}{NUAA}&Real&	87.33&	99.94&	88.01&	79.33&	90.09&	100.00&	95.87&	100.00\\
									 &Fake&	99.99&	99.99&	99.97&	99.95&	99.95&	0.00&	91.49&	100.00\\\hline
		\multirow{2}{*}{OULU-NPU}&Real&96.16&	99.96&	96.01&	95.39&	87.37&	99.92&	99.89&	98.69\\
							&Fake&	63.11&	43.56&	79.46&	72.95&	72.64&	0.10&	22.11&	99.10\\\hline
		\multirow{2}{*}{Replay-Attack}&Real&	99.00&	100.00&	90.13&	85.60&	16.53&	100.00&	89.03&	99.00\\
								&Fake&	91.77&	98.89&	88.51&	90.98&	99.47&	0.02&	85.12&	99.81\\\hline
		\multirow{2}{*}{CASIA-FASD}&Real&	91.40&	99.80&	63.50&	75.30&	58.30&	100.00&	97.10&	99.15\\
		   						 &Fake&	99.93&	99.87&	99.87&	99.22&	99.71&	0.00&	93.29&	99.56\\\hline
		\multirow{2}{*}{MSU-MFSD}&Real&	96.97&	100.00&	81.60&	91.02&	80.09&	100.00&	96.97&	100.00\\
		   						&Fake&	76.58&	82.35&	93.77&	92.01&	89.29&	0.00&	75.95&	100.00\\\hline
	\end{tabular}
	
	\raggedright
	\footnotesize Note: Sensetime$^{1}$ represents the algorithm version 2.2.5, Sensetime$^{2}$ represents the algorithm version 2.2.1, Megvii$^{1}$ represents the algorithm version 3.2.23, and Megvii$^{2}$ represents the algorithm version 3.1.98.
	\vspace{-2mm}
	\caption{Comparison of CLIPC8 detection rates with various algorithms on five datasets.\vspace{-3mm}}
	\label{table_comparison_public}
\end{table*}
\SubSection{Comparison of the CLIPC8 with Other Models} \label{sec_compare_othermodel}
In this section, we applied the CLIPC8 model to two different datasets, NUAA and Replay-Attack, and its results are compared with other models, as shown in Tab.~\ref{comparative_experiment_on_NUAA_dataset} ~\ref{comparative_experiment_on_RA_dataset}. For the NUAA dataset, ACC (Accuracy) is used as the evaluation metric, and for the RA (Replay-Attack) dataset, ACER (Average Classification Error Rate) is employed. The results show that our method achieved an ACC of 100.00\% on the NUAA dataset and an ACER of 0.595\% on the RA dataset, both of which indicate good performance.
\begin{table}[t]
\centering
\normalsize
\begin{tabular}{cc}
\hline
Models & ACC \\\hline
SPMT~\cite{26song2017face} & 97.90 \\
ND-CNN~\cite{27alotaibi2017deep} & 98.99 \\
MSNet~\cite{28zheng2021attention} & 99.50 \\
InceptionV3-FF~\cite{30yang2022face} & 99.96 \\
CLIPC8 &	100.00\\\hline
\end{tabular}
\vspace{-2mm}
\caption{Comparative experiment on NUAA dataset.\vspace{-3mm}}
\label{comparative_experiment_on_NUAA_dataset}
\end{table}
\begin{table}[t]
	\centering
	\normalsize
		\begin{tabular}{cc}
			\hline
			Models & ACER \\\hline
			YCbCr+HSV-LBP~\cite{31boulkenafet2015face} & 2.90 \\
			Vggface+LBP~\cite{32jiang2019deep} & 0.90 \\
			Patch-Depth Fusion CNN~\cite{33atoum2017face} & 0.72 \\
			TransFAS~\cite{09chaudhry2023transfas} & 0.63 \\
			CLIPC8 & 0.595 \\
			FARCNN~\cite{34chen2019cascade} &	100.00\\\hline
		\end{tabular}
	\vspace{-2mm}
	\caption{Comparative experiment on RA dataset.\vspace{-3mm}}
	\label{comparative_experiment_on_RA_dataset}
\end{table}
\SubSection{Ablation Study on Text Classification Information} \label{sec_ablation_experiment}
To analyze the impact of introducing new datasets and their image-text information on the image detection rate, we designed two ablation experiments. First, we tested the CLASS-8 dataset using the untrained ViT-L/14 weight model, and then used the CLIPC8 model, but set the eight text classification labels to numbers 1 to 8 during the image prediction phase. The results of the experiment are shown in Tab.~\ref{tab_ablation_onclipc8}. The results indicate that the untrained model only had a detection rate of 6.52\% for real human images, but a noticeably higher detection rate for other noun categories, demonstrating that the ViT-L/14 model itself has strong object recognition capabilities. However, when the classification text information was replaced with numbers 1 to 8, the CLIPC8 model was unable to effectively understand the connection between images and text, leading to a significant decrease in detection rate. In this experiment, we also found that the model tends to output fixed text classification, and this number represented \enquote{Identify Card Photo}, resulting in a 100\% detection rate for this category, but this does not indicate that the model correctly understood the relationship between image and text in the \enquote{Identify Card Photo} category.

\indent In summary, the ablation experiments based on the CLASS-8 dataset show that the introduction of textual description information from new datasets significantly improves the detection rate of face anti-spoofing. Accurate text classification labels help the model better capture the content information in the images.
\begin{table}[t]
	\centering
	\normalsize
	\begin{tabular}{ccc}
		\hline
		Classes & ViT-L/14 & CLIPC8*\\\hline
		(a) & 6.52&	0.00 \\
		(b) & 20.12&	100.00 \\
		(c) & 99.71&	0.00 \\
		(d) & 62.29&	0.00 \\
		(e) & 71.15&	0.46 \\
		(f) & 81.99&	6.90\\
		(g) & 15.56&	15.24\\
		(h) & 36.64&	0.00\\\hline
	\end{tabular}
	
	\raggedright
	\footnotesize Note: CLIPC8* represents that the classification text used in the CLIPC8 model is numbers 1-8.
	\vspace{-2mm}
	\caption{Ablation study of the CLIPC8 algorithm on the CLASS-8 dataset .\vspace{-3mm}}
	\label{tab_ablation_onclipc8}
\end{table}

\SubSection{Visual Analysis of the CLIPC8 Model} \label{sec_visual_onclip8}
This section focuses on analyzing the interpretability of the model, through two aspects of visualization analysis: first, the relationship between the cosine similarity of images and text, and second, the capability of the image encoder to extract feature information. These analyses demonstrate the model's ability to recognize the correlation between images and text and its effectiveness in liveness detection. Additionally, we compared the focus of the image encoder in the CLIPC8 and the original ViT-L/14 models on different local parts of the image in the attention matrix.

\subsubsection{Visual-language Saliency Heatmaps} \label{sec_visual_cosine_similarity}
\begin{figure}[htp]
\noindent 
\begin{subfigure}[b]{0.32\linewidth}
\includegraphics[width=\linewidth]{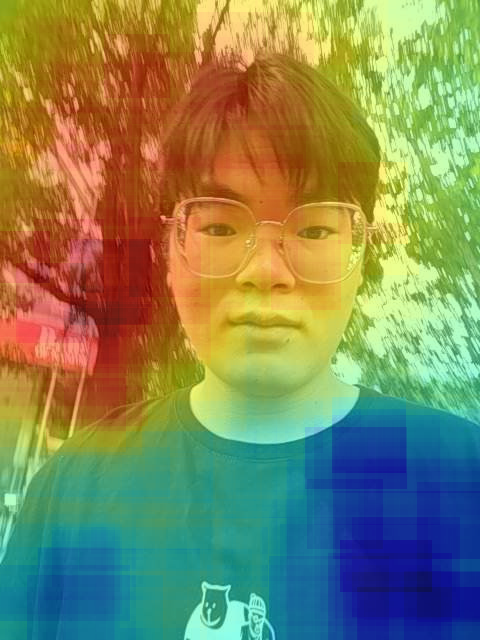}
\caption{}
\label{figure_similarity_res_for_image_text_pairs_a}
\end{subfigure}%
\hspace{2pt}
\begin{subfigure}[b]{0.32\linewidth}
\includegraphics[width=\linewidth]{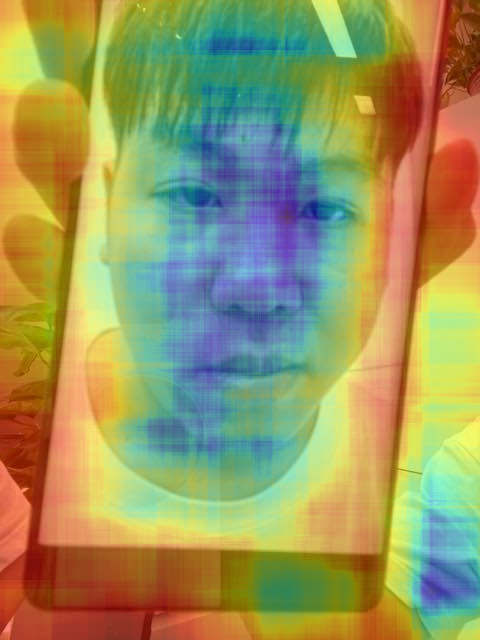}
\caption{}
\label{figure_similarity_res_for_image_text_pairs_b}
\end{subfigure}%
\hspace{2pt}
\begin{subfigure}[b]{0.32\linewidth}
\includegraphics[width=\linewidth]{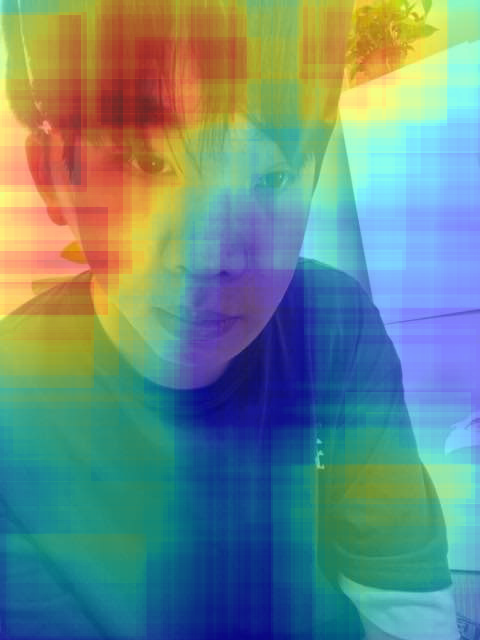}
\caption{}
\label{figure_similarity_res_for_image_text_pairs_c}
\end{subfigure}
\begin{subfigure}[b]{\linewidth}
\includegraphics[width=\linewidth]{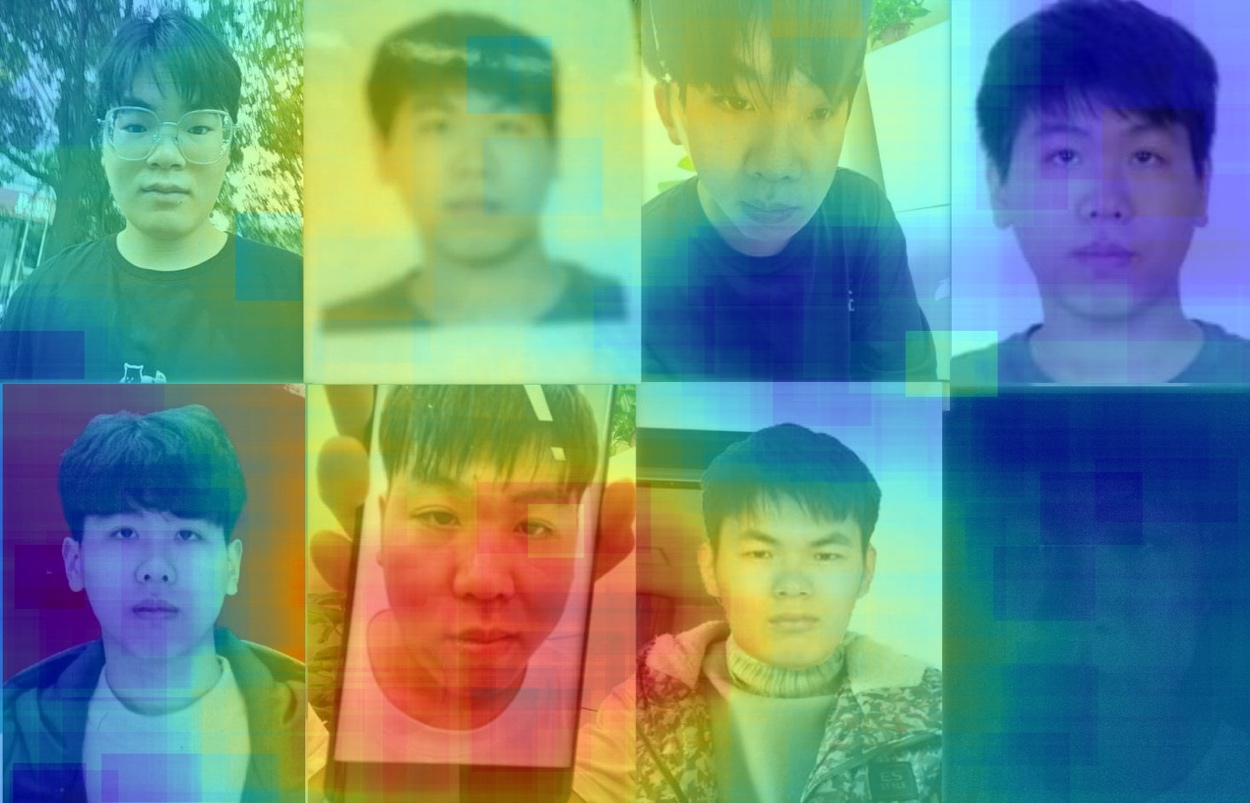}
\caption{}
\label{figure_similarity_res_for_image_text_pairs_d}
\end{subfigure}
\caption{Visualization of similarity results for image-text pairs}
\label{figure_similarity_res_for_image_text_pairs}
\end{figure}

\indent To better understand the details of how the proposed CLIPC8 model processes the correlation between image-text pairs, random sub-images from the original were cropped. The CLIPC8 model was used to calculate the cosine similarity between these sub-images and text descriptions, generating Visual-language Saliency Heatmaps. Overlaying these heatmaps on the original images provided an intuitive representation of the model's focus: low cosine similarity was mapped to deep blue, high similarity to bright red, and intermediate values were represented by a mix of colors. As shown in Fig.~\ref{figure_similarity_res_for_image_text_pairs}, in the Fig.~\ref{figure_similarity_res_for_image_text_pairs_a} the algorithm's attention is mainly concentrated on the upper part of the real human face, and in Fig.~\ref{figure_similarity_res_for_image_text_pairs_b} the recaptured photo attack using a smartphone displaying a face image, with the input text \enquote{mobile phone border} the attention was noticeably concentrated around the black edges of the smartphone. In the case of the face mask image Fig.~\ref{figure_similarity_res_for_image_text_pairs_c}, with the input text \enquote{mask} the algorithm's focus was predominantly on the upper-left corner where the mask was located. Fig.~\ref{figure_similarity_res_for_image_text_pairs_d} visualizes a composite image of eight samples from the CLASS-8 dataset, with the input text \enquote{smartphone} the algorithm accurately located the key part of the smartphone screen in the lower part of the composite image, but showing lower similarity for other components.

\subsubsection{Visualize Self Attention of CLIPC8} \label{sec_visual_self_attention}
\begin{figure}[htp]
	\centering
	\begin{subfigure}[t]{0.32\linewidth}
		\includegraphics[width=\linewidth]{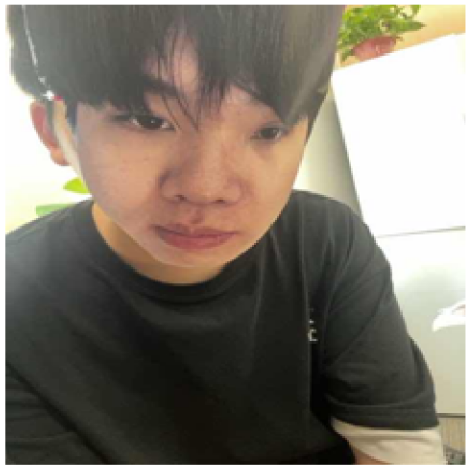}
		\label{figure_visual_self_attention_1}
	\end{subfigure}%
	\hfill
	\begin{subfigure}[t]{0.32\linewidth}
		\includegraphics[width=\linewidth]{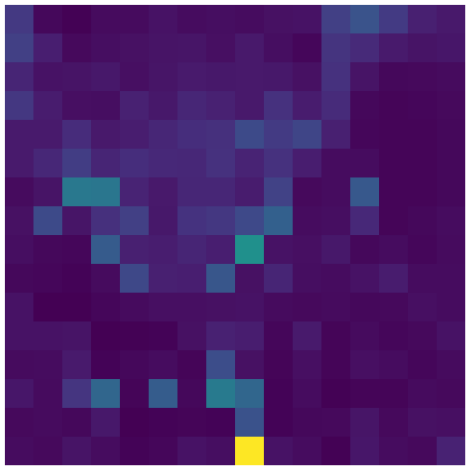}
		\caption{}
		\label{figure_visual_self_attention_2}
	\end{subfigure}%
	\hfill
	\begin{subfigure}[t]{0.32\linewidth}
		\includegraphics[width=\linewidth]{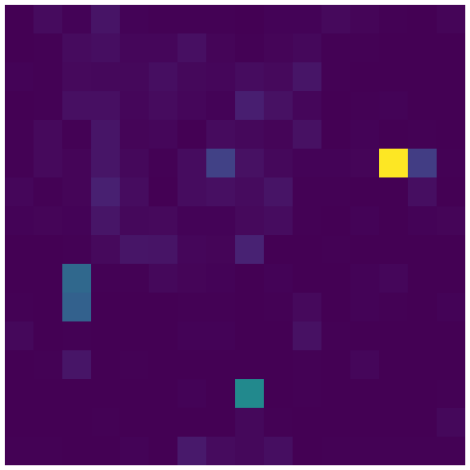}
		\label{figure_visual_self_attention_3}
	\end{subfigure}
	\begin{subfigure}[t]{0pt}
		\label{figure_visual_self_attention_a}
	\end{subfigure}
	\begin{subfigure}[t]{0.32\linewidth}
		\includegraphics[width=\linewidth]{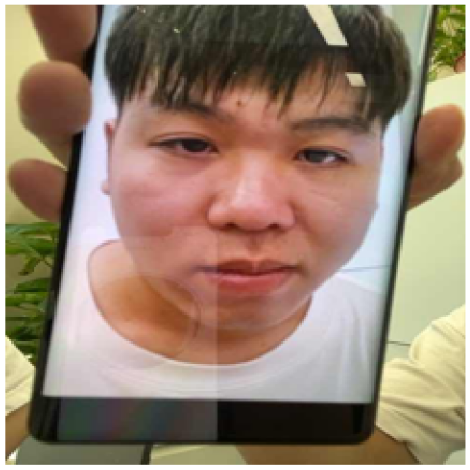}
		\label{figure_visual_self_attention_4}
	\end{subfigure}%
	\hfill
	\begin{subfigure}[t]{0.32\linewidth}
		\includegraphics[width=\linewidth]{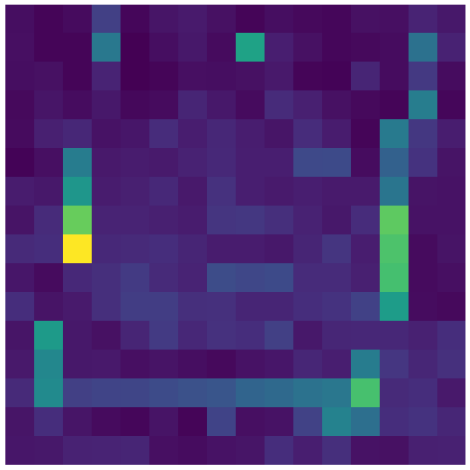}
		\caption{}
		\label{figure_visual_self_attention_5}
	\end{subfigure}%
	\hfill
	\begin{subfigure}[t]{0.32\linewidth}
		\includegraphics[width=\linewidth]{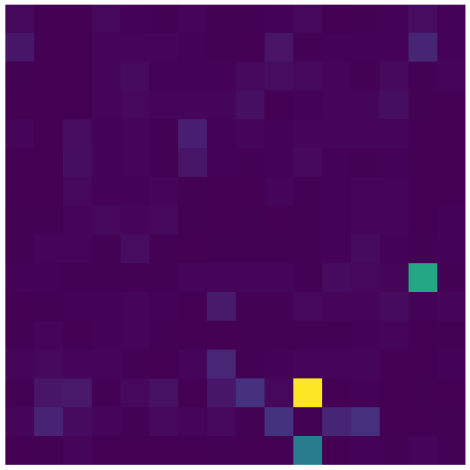}
		\label{figure_visual_self_attention_6}
	\end{subfigure}
	\begin{subfigure}[t]{0pt}
		\label{figure_visual_self_attention_b}
	\end{subfigure}
	\begin{subfigure}[t]{0.32\linewidth}
		\includegraphics[width=\linewidth]{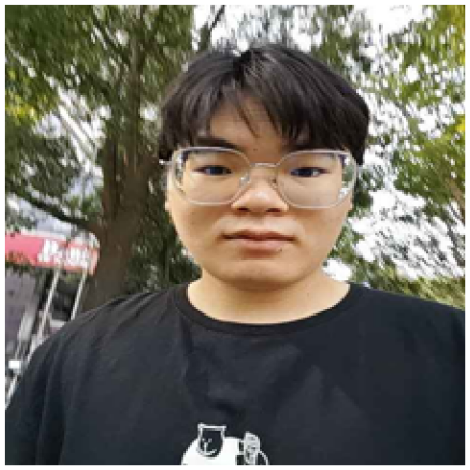}
		\label{figure_visual_self_attention_7}
	\end{subfigure}%
	\hfill
	\begin{subfigure}[t]{0.32\linewidth}
		\includegraphics[width=\linewidth]{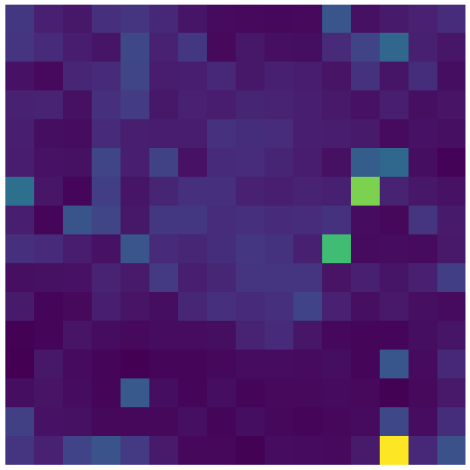}
		\caption{}
		\label{figure_visual_self_attention_8}
	\end{subfigure}%
	\hfill
	\begin{subfigure}[t]{0.32\linewidth}
		\includegraphics[width=\linewidth]{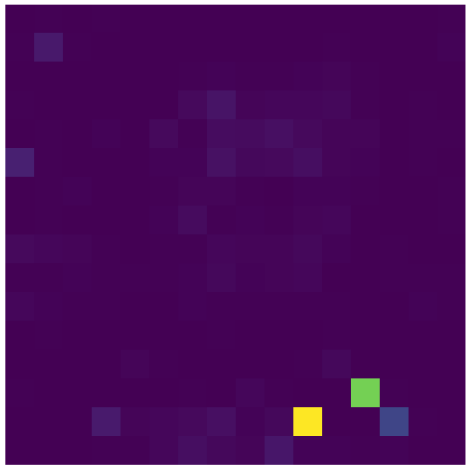}
		\label{figure_visual_self_attention_9}
	\end{subfigure}
	\begin{subfigure}[t]{0pt}
		\label{figure_visual_self_attention_c}
	\end{subfigure}
	\caption{Visualization of Self-Attention in the CLIPC8 Model}
\label{figure_visual_self_attention}
\end{figure}

\indent Fig.~\ref{figure_visual_self_attention} shows the attention distribution of the CLIPC8 model and the original ViT-L/14 model on the input image, using the output of the last layer of the self-attention layer of the image encoder in each model as the attention matrix. As shown in Fig.~\ref{figure_visual_self_attention}, the first column displays the original images, the second column shows the visualization of the corresponding images processed through the final layer of the CLIPC8 image encoder, and the third column depicts the visualization from the original ViT-L/14 model's image encoder.
In Fig.~\ref{figure_visual_self_attention}\subref{figure_visual_self_attention_2}, the CLIPC8 algorithm focused significantly on facial information in real human face images, showing a strong reaction. In the case of the recaptured photo attack Fig.~\ref{figure_visual_self_attention}\subref{figure_visual_self_attention_5}, the model focuses on the frame of the mobile screen.In the mask case, the attention matrix of the CLIPC8 model shows a strong reaction around the mask. Comparing with the original ViT-L/14 model, unrefined by the CLASS-8 dataset, the output from the final layer failed to capture key information in the images. These visual comparisons to some extent validate the theoretical significance and practical value of the algorithm and concepts proposed in this paper, demonstrating how the introduction of textual information and contrastive learning help the image encoder more accurately identify the features of live attack detection.

\SubSection{Limitation and Future Direction of the Proposed Method} \label{sec_limitation}
The method proposed in this article aims to solve the problem of insufficient generalization capability of commercial liveness detection algorithms in the financial field in specific scenarios. This problem stems from the limitations of scenarios and datasets. Our CLIPC8 training method relies on high-quality image-text pairs, which requires the construction of a large-scale annotated dataset. This dataset includes real photo and various types of attack images. In practical applications in the financial field, there may be an imbalance in the dataset categories, where the number of normal facial images far exceeds that of attack images. This could lead to model bias and overfitting. To address this issue, future work will focus on enhancing the data of minority categories and generating more attack training samples. Special attention will be paid to attack categories in optimizing the loss function, giving them greater weight. In the dataset construction process, to adopt an incremental approach, first using a small-scale dataset to train the benchmark model, and then performing preliminary classification on large-scale unlabeled data to reduce the workload of manual annotation.

The training method of this paper depends on image and text labels. In the selection of eight categories of text labels, we focus on the significant common features of each category of images. There is a problem with the generalization of descriptive text and whether to use Chinese or English for the labels. In future work, we can further refine the classification of normal and attack images, and initially generate text labels using an image-to-text description model, then use a text similarity classification model to preliminarily divide the dataset.

\Section{Conclusion}
This paper evaluated the detection rate of four commercialized private liveness detection algorithms in the financial field. The results show that current commercial liveness detection algorithms lack the capability to detect attacks in specific environments and have not achieved ideal results on the commonly used five major public live detection datasets. Therefore, we proposed a liveness detection method CLIPC8 based on image-text pair combined with contrastive learning and verified its good transfer effect on the five major public datasets. Compared with similar methods using Transformer for liveness detection, our method introduces text classification label information and the concept of contrastive learning into the live detection task, thereby improving the robustness and transferability of the algorithm on unseen datasets. The method of this paper complements the algorithm capability of commercialized private live detection algorithms in the financial field, improving the detection rate in special live attack scenarios.

However, there is still room for improvement in the test results of our method on the five public datasets under the Zero-shot testing method. We believe that as the CLASS-8 dataset continues to expand, especially with the addition of various types of attack samples and refined text category descriptions, the detection rate of the model will continue to rise.

{\small
\bibliographystyle{ieee_fullname}
\bibliography{00_dfd}
}
\end{document}